\def\year{2019}\relax
\documentclass[letterpaper]{article} %DO NOT CHANGE THIS
\usepackage{aaai19}  %Required
\usepackage{times}  %Required
\usepackage{helvet}  %Required
\usepackage{courier}  %Required
\usepackage{url}  %Required
\usepackage{graphicx}  %Required
\usepackage{amsmath}
\DeclareMathOperator{\sign}{sign}

\usepackage{subfigure}
\begin{document}
% The file aaai.sty is the style file for AAAI Press
% proceedings, working notes, and technical reports.
%
\title{Detection based Defense against Adversarial Examples from the Steganalysis Point of View}
\author{Jiayang Liu, Weiming Zhang, Yiwei Zhang, Dongdong Hou, Yujia Liu, Hongyue Zha and Nenghai Yu\\
CAS Key Laboratory of Electromagnetic Space Information\\
University of Science and Technology of China\\
Hefei, China\\
%2275 East Bayshore Road, Suite 160\\
%Palo Alto, California 94303\\
}
\maketitle
\begin{abstract}
Deep Neural Networks (DNNs) have recently led to significant improvements in many fields. However, DNNs are vulnerable to adversarial examples which are samples with imperceptible perturbations while dramatically misleading the DNNs. Moreover, adversarial examples can be used to perform an attack on various kinds of DNN based systems, even if the adversary has no access to the underlying model. Many defense methods have been proposed, such as obfuscating gradients of the networks or detecting adversarial examples. However it is proved out that these defense methods are not effective or cannot resist secondary adversarial attacks. In this paper, we point out that steganalysis can be applied to adversarial examples detection, and propose a method to enhance steganalysis features by estimating the probability of modifications caused by adversarial attacks. Experimental results show that the proposed method can accurately detect adversarial examples. Moreover, secondary adversarial attacks cannot be directly performed to our method because our method is not based on a neural network but based on high-dimensional artificial features and FLD (Fisher Linear Discriminant) ensemble.
\end{abstract}

\section{Introduction}

Deep Neural Networks (DNNs) have recently led to significant improvements in many fields, such as image classification \cite{russakovsky2015imagenet,he2016deep} and speech recognition \cite{amodei2016deep}. However, the generalization properties of the DNNs have been recently questioned because these machine learning models are vulnerable to adversarial examples \cite{szegedy2014intriguing}. An adversarial example is a slightly modified sample that is intended to cause an error output of the DNN based model. In the context of classification task, the adversarial example is crafted to force a model to classify it into a class different from the legitimate class. In addition, adversarial examples have cross-model generalization property \cite{Goodfellow2015Explaining}, so the attacker can even generate adversarial examples without the knowledge of the DNN. Adversarial attacks are divided into two types: targeted attack and untargeted attack. In targeted attack, the attacker generates adversarial examples which are misclassified by the classifier into a particular class. In untargeted attack, the attacker generates adversarial examples which are misclassified by the classifier into any class as long as it is different from the true class.

There are many studies which focus on methods of generating adversarial examples. Some attack methods are based on calculating the gradient of the network, such as Fast Gradient Sign Method (FGSM) \cite{Goodfellow2015Explaining}, Iterative Gradient Sign Method (IGSM) \cite{kurakin2016adversarial} and Jacobian Saliency Map Attack Method \cite{Papernot2016The}. While other methods are based on solving optimization problems, such as L-BFGS \cite{szegedy2014intriguing}, Deepfool \cite{Moosavidezfooli2015DeepFool} and Carlini \& Wagner (C\&W) attack \cite{Carlini2017Towards}.

Many defenses are proposed to mitigate adversarial examples against the above attacks. They make it harder for the adversary to craft adversarial examples using existing techniques or make the DNNs still give correct classifications on adversarial examples. These defenses are mainly divided into two categories.

One way is preprocessing the input image before classification, taking advantage of the spatial instability of adversarial examples. Defenders can perform some operations on the input image in spatial domain before giving the input image to a DNN, such as JPEG compression, scaling, adding noise, etc. Gu et al. \cite{gu2015towards} propose to use an autoencoder to remove adversarial perturbations from inputs.

The other way is to modify the network architecture, the optimization techniques or the training process. Goodfellow et al. \cite{Goodfellow2015Explaining} propose to augment the training set with adversarial examples to increase the model's robustness against a specific adversarial attack. However, this approach faces difficulties because the dimension of the images and features in networks means an unreasonable quantity of training data is required. Zheng et al. \cite{zheng2016improving} propose to append a stability term to the objective function to force the model to have similar outputs for normal images of the training set and their adversarial examples. This is different from data augmentation because it encourages the smoothness of the model output between original and adversarial samples. Defensive distillation \cite{papernot2016distillation} is another technique against certain adversarial attacks. This form of network can prevent the model from fitting too tightly to the original data. Unfortunately, most of these defenses are not very effective against adversarial examples in classification tasks.

Obfuscated gradients appear to be robust against adversarial attacks. Obfuscated gradients can be defined as a special case of gradient masking \cite{papernot2017practical}, in which the attackers cannot compute out the feasible gradient to generate adversarial examples. However, Athalye et al. \cite{athalye2018obfuscated} proposed attack techniques to overcome the obfuscated gradient based defenses.

Due to the difficulty of classifying adversarial examples correctly, recent work has turned to detecting them. Hendrycks \& Gimpel \cite{hendrycks2016early} use PCA to detect natural images from adversarial examples, finding that adversarial examples place a higher weight on the larger principal components than normal images. Li et al. \cite{li2017adversarial} apply PCA to the values after inner convolutional layers of the neural network, and use a cascade classifier to detect adversarial examples. Grosse et al. \cite{grosse2017statistical} propose a variant on adversarial re-training. They introduce a new class in the model solely for the adversarial examples, and train the network to detect adversarial examples. Gong et al. \cite{gong2017adversarial} construct a very similar defense technique as Grosse's defense. Metzen et al. \cite{metzen2017detecting} propose to add a detection subnetwork which observes the state of the original classification network, one can tell whether it has been presented with an adversarial example or not. Lu et al. \cite{Lu2017SafetyNet} detect adversarial examples by hypothesizing that adversarial examples produce different patterns of ReLU activations in networks than what is produced by normal images.

Unfortunately, Carlini \& Wagner perform experiments to prove that most of these detecting methods are only effective on image databases with small size or only several classes. Moreover, Grosse's, Gong's and Metzen's defenses use a second neural network to classify images as normal or adversarial. However, neural networks used for detecting adversarial examples can also be bypassed \cite{carlini2017adversarial}. In fact, given an adversarial method can fool the original neural network, Carlini et al. show that with a similar method we can also fool the extended network for detection, which we call secondary adversarial attacks.
%Li's defense failed against C\&W attack \cite{carlini2017adversarial}. Hendrycks's, Grosse's and Gong's defenses are only effective for MNIST. Metzen's defense only argues robustness against CIFAR-10.
%However, the Hendrycks defense is only effective for MNIST. Unfortunately, Carlini \& Wagner performed experiments to prove that Li's defense failed against C\&W attack \cite{carlini2017adversarial}. However, Carlini \& Wagner re-implemented these two defenses and found that they are only effective for MNIST \cite{carlini2017adversarial}. However, this defense only argues robustness against CIFAR-10.

In this paper we propose to detect adversarial examples from the view of steganalysis \cite{pevny2010steganalysis} which is the technology for detecting steganography. In fact, Goodfellow et al. \cite{Goodfellow2015Explaining} have provided the insight on one essence of adversarial examples such that ``the adversarial attack can be treated as a sort of accidental steganography''. Furthermore, we propose a method to enhance steganalysis features by estimating the probability of modifications caused by adversarial attacks. Experimental results show that the proposed method can accurately detect adversarial examples. Moreover, secondary adversarial attacks cannot be directly performed to our method because our method is not based on a neural network but based on high-dimensional artificial features and FLD (Fisher Linear Discriminant) ensemble.

\section{Related Work}
\subsection{Adversarial Attacks}

\subsubsection{Fast Gradient Sign Method}
Goodfellow et al. \cite{Goodfellow2015Explaining} propose the Fast Gradient Sign Method (FGSM) for generating adversarial examples. This method uses the derivative of the loss function of the model pertaining to the input feature vector. Given the input image $X$, FGSM is to perturb the gradient direction of each feature by the gradient. Then the classification result of the input image will be changed. For a neural network with cross-entropy cost function $J(X,y)$ where $X$ is the input image and $y_{t}$ is the target class for the input image, the adversarial example is generated as\begin{equation}X^{adv}=X-\epsilon{\sign(\nabla_XJ(X,y_{t}))}\end{equation}where $\epsilon$ is a parameter to determine the perturbation size.

\subsubsection{Iterative Gradient Sign Method}
The Iterative Gradient Sign Method (IGSM) is the iterative version of FGSM. This method applies FGSM many times with small perturbation size instead of applying adversarial noise with one large perturbation size. The adversarial example of the iterative gradient sign method is generated as\begin{equation}\begin{split}X_0^{adv}&=X,\\X_{N+1}^{adv}=Clip_{X,\epsilon}\{X_N^{adv}-&\alpha{\sign(\nabla_XJ(X_N^{adv},y_{t}))}\}\end{split}\end{equation}where $Clip_{X,\epsilon}\{X'\}$ represents a clipping of the values of the adversarial example. So the results are within $\epsilon$-neighbourhood of the input image $X$. This attack is more powerful because the attacker can control how far the adversarial example past the classification boundary. It was demonstrated that the attack of IGSM was better than FGSM on ImageNet-1000 \cite{kurakin2016adversarial}.

\subsubsection{Deepfool}
Deepfool is an untargeted attack method to generate an adversarial example by iteratively perturbing an image \cite{Moosavidezfooli2015DeepFool}. This method explores the nearest decision boundary. The image is modified a little to reach the boundary in each iteration. The algorithm stops once the modified image changes the classification of the network.

\subsubsection{Carlini \& Wagner Method}
This method is named after its authors \cite{Carlini2017Towards}. The attack can be targeted or untargeted, and has three metrics to measure its distortion (${l_0}$ norm, ${l_2}$ norm and ${l_\infty }$ norm). The authors point out that the untargeted ${l_2}$ norm version has the best performance. It generates adversarial examples by solving the following optimization problem:\begin{equation}\begin{split}\mathop {{\rm{minimize}}}\limits_\delta  {\rm{  }}{\left\| \delta  \right\|_2} &+ c \cdot f(x + \delta )\\s.t. \quad {\rm{  }}x +& \delta  \in {\left[ {0,1} \right]^n}\end{split}\end{equation}

This attack is to look for the smallest perturbation measured by ${l_2}$ norm and make the network classify the image incorrectly at the same time. $c$ is a hyperparameter to balance the two parts of equation $\left( 3 \right)$. The best way to choose $c$ is to use the smallest value of $c$ for which the resulting solution $x + \delta$ has $f\left( {x + \delta } \right) \le 0$. $f(x)$ is the loss function to measure the distance between the input image and the adversarial image. $f(x)$ is defined as: \begin{equation}f(x) = \max (Z{(x)_{true}} - {\max _{i \ne true}}\{ Z{(x)_i}\} , - \kappa )\end{equation}

$Z(x)$ is the pre-softmax classification result vector. $\kappa$ is a hyper-parameter called confidence. Higher confidence encourages the attack to search for adversarial examples that are stronger in classification confidence. High-confidence attacks often have larger perturbations and better transferability to other models. The C\&W method is a strong attack which is difficult to defend.

\subsection{Robustness Based Defense}
Robustness based defense aims at classifying adversarial examples correctly. There are many methods to achieve robustness based defense. Adversarial training is to train a better network by using a mixture of normal and adversarial examples in the training set for data augmentation, which we refer to as adversarial training \cite{Goodfellow2015Explaining}. Preprocessing the input images is to perform some operations to remove adversarial perturbations, such as principal component analysis (PCA) \cite{bhagoji2017dimensionality}, JPEG compression \cite{Das2017Keeping}, adding noise, cropping, rotating and so on. Defensive distillation hides the gradient between the pre-softmax layer and softmax outputs by leveraging distillation training techniques \cite{papernot2016distillation}. Obfuscated gradients make the attackers hard to compute out the feasible gradient to generate adversarial examples \cite{athalye2018obfuscated}.

\subsection{Detection Based Defense}
Detection based defense aims at distinguishing normal images and adversarial examples.

Hendrycks \& Gimpel \cite{hendrycks2016early} use PCA to detect natural images from adversarial examples, finding that adversarial examples place a higher weight on the larger principal components than normal images. However, the Hendrycks defense is only effective for MNIST.

Li et al. \cite{li2017adversarial} apply PCA to the values after inner convolutional layers of the neural network, and use a cascade classifier to detect adversarial examples. Specifically, they propose building a cascade classifier that accepts the input as natural only if all classifiers accept the input, but rejects it if any do. However, Carlini \& Wagner perform experiments to prove that Li's defense fails against the C\&W attack \cite{carlini2017adversarial}.

Grosse et al. \cite{grosse2017statistical} propose a variant on adversarial re-training. Instead of attempting to classify the adversarial examples correctly, they introduce the $\left( {N + 1} \right)th$ class, solely for adversarial examples, and train the network to detect adversarial examples. Gong et al. \cite{gong2017adversarial} construct a very similar defense technique. However, Carlini \& Wagner re-implement these two defenses and find that they are only effective for MNIST \cite{carlini2017adversarial}.

Metzen et al. \cite{metzen2017detecting} detect adversarial examples by looking at the inner convolutional layers of the network. They augment the classification neural network with a detection neural network that takes its input from various intermediate layers of the classification network. However, this defense is only effective against CIFAR-10.

Lu et al. \cite{Lu2017SafetyNet} hypothesize that adversarial examples produce different patterns of ReLU activations in networks than what is produced by normal images. Based on this hypothesis, they propose the Radial Basis Function SVM (RBF-SVM) classifier which takes advantage of discrete codes computed by the late stage ReLUs of the network to detect adversarial examples on CIFAR-10 and ImageNet-1000.

For practical applications, we can deploy detection based defense combing with robustness based defense. First of all, we use detection based defense to detect the input image. If it is a normal image, we will directly feed it to the original DNN. Otherwise we can take advantage of robustness based defense to mitigate adversarial examples.

\section{Proposed Method}
Both adversarial attacks and steganography make perturbations on the pixel values, which alter the dependence between pixels. However, steganalysis can effectively detect modifications caused by steganography by modeling the dependence between adjacent pixels in natural images. So we can also take advantage of steganalysis to identify deviations due to adversarial attacks.

%We recommend using two models to extract steganalysis features. One is the low-dimensional model called SPAM with 686 features. The other is the high-dimensional model called SRM with 34671 features.

Assuming that we have known the attacking method used by the attacker, we construct a detector to detect whether the input image is an adversarial example or not. In practice, we don't know the method used by the attacker, but we can deploy a series of detectors trained for various mainstream adversarial attacks. Our detection method exploits the fact that the perturbation of pixel values by adversarial attack alters the dependence between pixels. By modeling the differences between adjacent pixels in natural images, we can identify deviations due to adversarial attacks. In the beginning, we use a filter to suppress the content of the input image. Dependence between adjacent pixels of the filtered image is modeled as a higher order Markov chain \cite{sullivan2005steganalysis}. Then the transition probability matrix is used as a vector feature for a feature based detector implemented using machine learning algorithms. %Moreover, we propose a framework of defense mechanism against adversarial examples. First of all, we use our detector to detect the input image. If it is a normal image, we will directly feed it to the neural network to get the correct classification. Otherwise we can take advantage of the above defenses to mitigate adversarial examples.

We recommend two kinds of steganalysis feature sets for detecting adversarial examples: one is the low-dimensional model SPAM with 686 features \cite{pevny2010steganalysis}; the other is the high-dimensional model Spatial Rich Model (SRM) with 34671 features \cite{fridrich2012rich}.

%\begin{figure}[ht]
%  \centering
%  \includegraphics[width=0.5\textwidth]{framework.pdf}
%  \caption{Framework of defense mechanism.}
%  \label{framework}
%\end{figure}

\subsection{Features Extraction}
\subsubsection{SPAM}
SPAM is described as follows. First, the model calculates the transition probabilities between pixels in eight directions $\left\{ { \leftarrow , \to , \downarrow , \uparrow , \nwarrow , \searrow , \swarrow , \nearrow } \right\}$ in the spatial domain. The differences and the transition probability are always computed along the same direction. For example, the horizontal direction from left-to-right differences are calculated by $A_{i,j}^ \to  = {X_{i,j}} - {X_{i,j + 1}}$, where $X$ is an image with size of $m \times n$, and ${X_{i,j}}$ is the pixel at the position $\left( {i,j} \right)$ for $i \in \left\{ {1, \ldots ,m} \right\}$, $j \in \left\{ {1, \ldots ,n - 1} \right\}$. Second, to model pixel dependence along the eight directions, a Markov chain is used between pairs of differences (first order chain) or triplets (second order chain). The first-order detecting features, ${F^{1st}}$, model the difference arrays $A$ by a first-order Markov process. For the horizontal direction, this leads to\begin{equation}M_{x,y}^ \to  = Pr(A_{i,j + 1}^ \to  = x|A_{i,j}^ \to  = y)\end{equation}where $x,y \in \{  - T, \ldots ,T\} $. The second-order detecting features, ${F^{2nd}}$, model the difference arrays $A$ by a second-order Markov process. For the horizontal direction, this leads to\begin{equation}M_{x,y,z}^ \to  = P(A_{i,j + 2}^ \to  = x|A_{i,j + 1}^ \to  = y,A_{i,j}^ \to  = z)\end{equation}where $x,y,z \in \{  - T, \ldots ,T\} $. For dimensionality reduction of the transition probability matrix, only differences within a limited range are considered. Thus, the transition probability matrix is calculated just for pairs within $[-T,T]$. We separately average the horizontal and vertical matrices and then the diagonal matrices to form the final feature sets, ${F^{1st}}$, ${F^{2nd}}$. The expression of the average sample Markov transition probability matrices is\begin{equation}\begin{split}{F_{1, \ldots ,k}} &= ({M^ \to } + {M^ \leftarrow } + {M^ \uparrow } + {M^ \downarrow })/4\\{F_{k + 1, \ldots ,2k}} &= ({M^ \nearrow } + {M^ \swarrow } + {M^ \searrow } + {M^ \nwarrow })/4\end{split}\end{equation}where $k = {(2T + 1)^2}$ for the first-order detecting features and $k = {(2T + 1)^3}$ for the second-order detecting features. We can see that the order of Markov model and the range of differences $T$ control the dimensionality of our detecting model. We use $T=3$ for second order, resulting in $2k = 686$ features \cite{pevny2010steganalysis}.

\subsubsection{Spatial Rich Model}
%We take advantage of Spatial Rich Model (SRM) to extract detecting features. SRM is the state-of-the-art high-dimensional model in steganalysis \cite{fridrich2012rich}. SRM calculates the co-occurrence matrix for four neighboring pixels with various submodels. SRM deals with every pixel within an image equally in feature extraction. Considering the impact of the Modification Probability Map, we use maxSRM to detect adversarial examples. The maxSRM is built in the same manner as the SRM but the process of forming the co-occurrence matrices is modified to consider the modification  probabilities \cite{denemark2014selection}.

Spatial Rich Model (SRM) can be viewed as an extended version of SPAM by extracting residuals from images \cite{fridrich2012rich}. A residual is an estimate of the image noise component obtained by subtracting from each pixel its estimate obtained using a pixel predictor from the pixel's immediate neighborhood. SRM uses 45 different pixel predictors of two different types: linear and non-linear. Each linear predictor is a shift-invariant finite-impulse response filter described by a kernel matrix ${K^{\left( {pred} \right)}}$. The residual $Z = \left( {{z_{kl}}} \right)$ is a matrix of the same dimension as $X$: \begin{equation}Z = {K^{\left( {pred} \right)}} * X - X \buildrel \Delta \over = K * X\end{equation}where the symbol $*$ denotes the convolution with $X$ mirror-padded so that $K * X$ has the same dimension as $X$.

An example of a simple linear residual is ${z_{ij}} = {X_{i,j + 1}} - {X_{i,j}}$, which is the difference between a pair of horizontally neighboring pixels. In this case, the residual kernel is $K = \left( {\begin{array}{*{20}{c}}{ - 1}&1 \end{array}} \right)$, which means that the predictor estimates the pixel value as its horizontally adjacent pixel.

All non-linear predictors in the SRM are obtained by taking the minimum or maximum of up to five residuals obtained using linear predictors. For example, one can predict pixel ${X_{i,j}}$ from its horizontal or vertical neighbors, obtaining thus one horizontal and one vertical residual ${Z^{\left( h \right)}} = \left( {z_{ij}^h} \right)$, ${Z^{\left( v \right)}} = \left( {z_{ij}^v} \right)$:\begin{equation}z_{ij}^{\left( h \right)} = {X_{i,j + 1}} - {X_{i,j}}\end{equation}\begin{equation}z_{ij}^{\left( v \right)} = {X_{i + 1,j}} - {X_{i,j}}\end{equation}

Using these two residuals, one can compute two nonlinear ``minmax'' residuals as:\begin{equation}z_{ij}^{\left( {\min } \right)} = \min \left\{ {z_{ij}^{\left( h \right)},z_{ij}^{\left( v \right)}} \right\}\end{equation}\begin{equation}z_{ij}^{\left( {\max } \right)} = \max \left\{ {z_{ij}^{\left( h \right)},z_{ij}^{\left( v \right)}} \right\}\end{equation}

After that, quantize $Z$ with a quantizer ${Q_{ - T,T}}$ with centroids ${Q_{ - T,T}} = \left\{ { - Tq,\left( { - T + 1} \right)q, \ldots ,Tq} \right\}$, where $T > 0$ is an integer threshold
and $q>0$ is a quantization step:\begin{equation}{r_{ij}} \buildrel \Delta \over = {Q_{ - T,T}}\left( {{z_{ij}}} \right),\forall i,j\end{equation}

The next step in forming the SRM feature vector involves computing a co-occurrence matrix of fourth order, ${C^{\left( {SRM} \right)}} \in Q_{ - T,T}^4$, from four (horizontally and vertically) neighboring values of the quantized residual ${r_{ij}}$ from the entire image:\begin{equation}C_{{d_0}{d_1}{d_2}{d_3}}^{SRM} = \sum\limits_{i,j = 1}^{m,n - 3} {\left[ {{r_{i,j}} = {d_k},\forall k = 0, \ldots ,3} \right]} \end{equation}where $\left[ B \right]$ is the Iverson bracket, which is equal to 1 when the statement $B$ is true and to 0 when it is false. The union of all co-occurrence matrices, including their differently quantized versions, has a total dimension of 34671.

%\subsection{Estimation of Modification Probability Map}
\subsection{Features Enhancement}
The above methods of extracting steganalysis features do not consider the location of modified pixels caused by adversarial attacks. Obviously, the accuracy of detection will be improved if we assign larger weight to the features of modified location. Although we cannot obtain the accurate modified location, we can estimate the relative modification probability of each pixel. In order to further improve the accuracy of detection, we propose to enhance steganalysis features by estimating the probability of modifications caused by adversarial attacks.

We take advantage of the gradient amplitude to estimate the modification probability because the pixels with larger gradient amplitude have larger probability to be modified. Assume that the neural network divides images into $N$ categories. Although we cannot know which target class will be selected by the attacker, we can randomly select $L$ categories to generate $L$ targeted adversarial examples and then estimate the modification probability of each pixel according to these targeted adversarial examples. So we take the $ith$ $\left( {1 \le i \le L} \right)$ class as the target class to calculate the gradient of the input image $X$. We refer to the matrix of all pixels' modification probabilities as Modification Probability Map (MPM). Note that these targeted adversarial examples generated by us are only used to estimate MPM which can be used to detect untargeted attacks.

For FGSM and IGSM, when generating the adversarial example of the target class ${y_i}$ for the input image, we save absolute values of the gradient of each pixel $\left| {{\nabla _X}J(X,{y_i})} \right|$, and then normalize them to obtain the gradient map ${f_{nor}}\left( {\left| {{\nabla _X}J(X,{y_i})} \right|} \right)$ where ${f_{nor}}\left( {} \right)$ is the function to normalize all elements in the matrix to $(0,1)$. Finally, calculate the mean value of the gradient maps of $L$ adversarial examples to get MPM $P$:\begin{equation}P = \frac{1}{L}\sum\limits_{i = 1}^L {{f_{nor}}\left( {\left| {{\nabla _X}J(X,{y_i})} \right|} \right)}\end{equation}

$P$ is a $m \times n$ matrix in which the $\left( {i,j} \right)th$ element ${P_{i,j}}$ is the modification probability of the $\left( {i,j} \right)th$ pixel ${X_{i,j}}$.

For C\&W which does not generate adversarial examples by gradient, the estimation of MPM starts by computing the difference array ${D_i}$ between the normal image $X$ and the adversarial example $X_i^{adv}$:\begin{equation}{D_i} = X_i^{adv} - X\end{equation}Then save the  absolute values of all elements in the difference array ${D_i}$ and normalize them to obtain the difference map ${f_{nor}}\left( {\left| {{D_i}} \right|} \right)$. Finally, calculate the mean value of the difference maps of $L$ adversarial examples to get MPM:\begin{equation}P = \frac{1}{L}\sum\limits_{i = 1}^L {{f_{nor}}\left( {\left| {{D_i}} \right|} \right)}\end{equation}

For Deepfool which can only generate untargeted adversarial examples, we estimate MPM by computing the difference array $D$ between the normal image $X$ and the adversarial example ${X^{adv}}$:\begin{equation}D = {X^{adv}} - X\end{equation}Then save the absolute values of all elements in the difference array $D$ and normalize them to obtain MPM:\begin{equation}P = {f_{nor}}\left( {\left| D \right|} \right)\end{equation}

\begin{figure*}[htbp]
\centering
\subfigure[Normal image]{\includegraphics[width=0.24\textwidth]{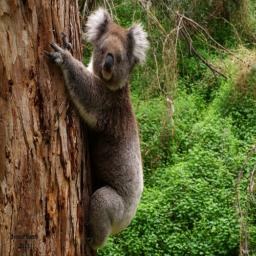}\label{normal image}}
\subfigure[MPM of normal image]{\includegraphics[width=0.24\textwidth]{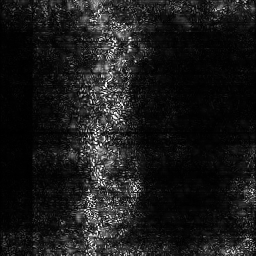}\label{normal map}}
\subfigure[Adversarial image]{\includegraphics[width=0.24\textwidth]{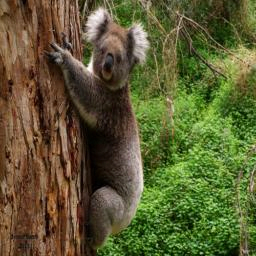}\label{adversarial image}}
\subfigure[MPM of adversarial image]{\includegraphics[width=0.24\textwidth]{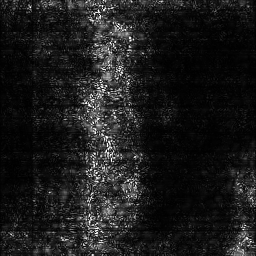}\label{adversarial map}}\\
\caption{Illustrations of a normal image, an adversarial image and their MPM.}
\label{Illustrations}
\end{figure*}

The above description is the estimation of MPM based on normal images. In practice, the detector may receive an adversarial example. The results of our experiments show that the MPM of one normal image and its adversarial image is quite similar. Figure \ref{Illustrations} shows an example of a normal image, an adversarial image and their MPM (normalized to $(0,255)$ to show more clearly).

\subsubsection{Enhanced SPAM}
Considering the impact of MPM, Enhanced SPAM (ESPAM) is proposed. The difference between SPAM and ESPAM is that we construct a new Markov transition probability based on MPM. For example, in the horizontal direction, the Markov transition probability $M_{x,y}^ \to$ is related to the pixel $X_{i,j}$, $X_{i,j+1}$ and $X_{i,j+2}$. So we calculate the new Markov transition probability $M'^ \to_{x,y}$ in this way: \begin{equation}M'^ \to_{x,y}  = M_{x,y}^ \to  \cdot {P_{i,j}} \cdot {P_{i,j + 1}} \cdot {P_{i,j + 2}}\end{equation}Similarly, for the second-order detecting features, the new Markov transition probability $M'^ \to_{x,y,z}$ is \begin{equation}M'^ \to_{x,y,z}  = M_{x,y,z}^ \to  \cdot {P_{i,j}} \cdot {P_{i,j + 1}} \cdot {P_{i,j + 2}} \cdot {P_{i,j + 3}}\end{equation}Then the expression of the average sample Markov transition probability matrices is\begin{equation}\begin{split}{F_{1, \ldots ,k}} &= ({M'^ \to } + {M'^ \leftarrow } + {M'^ \uparrow } + {M'^ \downarrow })/4\\{F_{k + 1, \ldots ,2k}} &= ({M'^ \nearrow } + {M'^ \swarrow } + {M'^ \searrow } + {M'^ \nwarrow })/4\end{split}\end{equation}where $k = {(2T + 1)^2}$ for the first-order detecting features and $k = {(2T + 1)^3}$ for the second-order detecting features. ESPAM has the same dimensionality as SPAM, which is 686.
\subsubsection{Enhanced SRM}
The Enhanced Spatial Rich Model (ESRM) is built in the same manner as the SRM but the process of forming the co-occurrence matrices is modified to consider the impact of MPM:\begin{equation}C_{{d_0}{d_1}{d_2}{d_3}}^{ESRM} = \sum\limits_{i,j = 1}^{m,n - 3} {\mathop {\max }\limits_{k = 0, \ldots ,3} {P_{i,j + k}}\left[ {{r_{i,j}} = {d_k},\forall k = 0, \ldots ,3} \right]} \end{equation}

Above, ${C^{\left( {ESRM} \right)}}$ denotes the enhanced version of the co-occurrence ${C^{\left( {SRM} \right)}}$. In other words, instead of increasing the corresponding co-occurrence bin by 1, the maximum of the modification probabilities taken across the four residuals is added to the bin \cite{denemark2014selection}. Thus, if a group has four pixels with small modification probabilities, it has smaller effect on the co-occurrence values than the group with at least one pixel likely to be changed. The rest of the process of forming ESRM stays exactly the same with SRM. ESRM has the same dimensionality as SRM, which is 34671.
\subsection{Training Detector}
The construction of our detectors based on features relies on pattern-recognition classifiers. The detectors are trained as binary classifiers implemented using the FLD ensemble \cite{kodovsky2012ensemble} with default settings. The ensemble by default minimizes the total classification error probability under equal priors. The random subspace dimensionality and the number of base learners is found by minimizing the out-of-bag estimate of the testing error on bootstrap samples of the training set as it is an unbiased estimate of the testing error on unseen data \cite{breiman1996bagging}.

\section{Experimental Results}
We construct the detectors by modeling the differences between adjacent pixels in natural images. Therefore, our method can not achieve very good performance on MNIST and CIFAR-10 because the size of the image is too small. However, it has good performance on ImageNet-1000. Previous work showed that untargeted attack is easier to succeed, results in smaller perturbations, and transfers better to different models. So we detect untargeted adversarial examples to see the performance of our method.

\begin{table}[htbp]
\caption{Detection accuracy of normal images and their adversarial images generated by FGSM.}
\centering
\begin{tabular}[h]{c | c | c | c | c }
\hline
SPAM    &$\epsilon=2$    &$\epsilon=4$     &$\epsilon=6$     &$\epsilon=8$  \\ \hline
normal images	     & 0.9488	& 0.9570  & 0.9651  & 0.9713  \\ \hline
adversarial images   & 0.9432   & 0.9559  & 0.9628  & 0.9709  \\ \hline
ESPAM    &$\epsilon=2$    &$\epsilon=4$     &$\epsilon=6$     &$\epsilon=8$  \\ \hline
normal images	     & 0.9725	& 0.9758  & 0.9812  & 0.9868  \\ \hline
adversarial images   & 0.9704   & 0.9719  & 0.9751  & 0.9806  \\ \hline
SRM    &$\epsilon=2$    &$\epsilon=4$     &$\epsilon=6$     &$\epsilon=8$  \\ \hline
normal images	     & 0.9757	& 0.9814  & 0.9831  & 0.9887  \\ \hline
adversarial images   & 0.9785   & 0.9822  & 0.9861  & 0.9903   \\ \hline
ESRM    &$\epsilon=2$    &$\epsilon=4$     &$\epsilon=6$     &$\epsilon=8$  \\ \hline
normal images	     & 0.9809	& 0.9839  & 0.9900  & 0.9931  \\ \hline
adversarial images   & 0.9811   & 0.9866  & 0.9905  & 0.9938  \\ \hline
RBF-SVM   &$\epsilon=2$    &$\epsilon=4$     &$\epsilon=6$     &$\epsilon=8$  \\ \hline
normal images	     & 0.8340	& 0.8913  & 0.9305  & 0.9487  \\ \hline
adversarial images	 & 0.8258   & 0.8936  & 0.9243  & 0.9541\\ \hline
\end{tabular}
\label{untarget fast vgg}
\end{table}

\begin{table}[htbp]
\caption{Detection accuracy of normal images and their adversarial images generated by IGSM.}
\centering
\begin{tabular}[h]{c | c | c | c | c }
\hline
SPAM    &$\epsilon=2$    &$\epsilon=4$     &$\epsilon=6$     &$\epsilon=8$  \\ \hline
normal images	     & 0.9402	& 0.9485  & 0.9559  & 0.9606  \\ \hline
adversarial images   & 0.9411   & 0.9474  & 0.9545  & 0.9601  \\ \hline
ESPAM    &$\epsilon=2$    &$\epsilon=4$     &$\epsilon=6$     &$\epsilon=8$  \\ \hline
normal images	     & 0.9708	& 0.9737  & 0.9749  & 0.9760  \\ \hline
adversarial images   & 0.9638   & 0.9675  & 0.9725  & 0.9745  \\ \hline
SRM    &$\epsilon=2$    &$\epsilon=4$     &$\epsilon=6$     &$\epsilon=8$  \\ \hline
normal images	     & 0.9667	& 0.9706  & 0.9753  & 0.9802  \\ \hline
adversarial images   & 0.9697   & 0.9724  & 0.9762  & 0.9812  \\ \hline
ESRM    &$\epsilon=2$    &$\epsilon=4$     &$\epsilon=6$     &$\epsilon=8$  \\ \hline
normal images	     & 0.9712	& 0.9754  & 0.9811  & 0.9878  \\ \hline
adversarial images   & 0.9716   & 0.9767  & 0.9820  & 0.9879  \\ \hline
RBF-SVM   &$\epsilon=2$    &$\epsilon=4$     &$\epsilon=6$     &$\epsilon=8$  \\ \hline
normal images	     & 0.7749	& 0.8660  & 0.9145  & 0.9362  \\ \hline
adversarial images	 & 0.7975   & 0.8752  & 0.9072  & 0.9330\\ \hline	
\end{tabular}
\label{untarget iterative vgg}
\end{table}

\begin{table}[htp]
\caption{Detection accuracy of normal images and their adversarial images generated by Deepfool.}
\centering
\begin{tabular}[h]{c | c | c }
\hline
   &normal images  &adversarial images \\ \hline
SPAM			         & 0.8553  & 0.8481\\ \hline
ESPAM			         & 0.8870  & 0.8629\\ \hline
SRM			         & 0.9445  & 0.9491\\ \hline
ESRM			     & 0.9498  & 0.9527\\ \hline
RBF-SVM              & 0.5838  & 0.6012      \\ \hline
\end{tabular}
\label{untarget deepfool vgg}
\end{table}

\begin{table}[htbp]
\caption{Detection accuracy of normal images and their adversarial images generated by C\&W.}
\centering
\begin{tabular}[h]{c | c | c }
\hline
   &normal images  &adversarial images \\ \hline
SPAM			         & 0.6957  & 0.6778\\ \hline
ESPAM			         & 0.8025  & 0.8296\\ \hline
SRM			         & 0.8814  & 0.9092\\ \hline
ESRM			     & 0.9233  & 0.9341\\ \hline
RBF-SVM			     & 0.5332  & 0.5187      \\ \hline
\end{tabular}
\label{untarget carlini vgg}
\end{table}

We test our detecting method against untargeted attacks by FGSM, IGSM, Deepfool and C\&W. Our experiments are performed on 40000 images randomly selected from ImageNet-1000 (ILSVRC-2016) using a pretrained VGG-16 model \cite{Simonyan2014Very} as classification network which is evaluated with top-1 accuracy. This results in a train set of 25000 images, a validation set of 5000 images, and a test set of 10000 images. The values of pixels per color channel of these 40000 images range from 0 to 255. For IGSM, we use $\alpha=1$ to ensure that we change each pixel by 1 on each step and $\epsilon \le 8$ where $\epsilon$ is a parameter to determine the perturbation size. For Deepfool, we apply the ${l_2}$ norm version. For C\&W, we set $\kappa=0$. In the process of estimating MPM, we set $L=100$.

At first, the 40000 images from ImageNet-1000 are classified by the network to obtain their true labels. Then we use these 40000 images to generate 40000 adversarial images as adversarial samples of our experiments. To prove that MPM is effective when detecting adversarial examples, we perform comparative experiments. We construct two pairs of detectors: SPAM and ESPAM, SRM and ESRM. The only difference between each pair is one detector with MPM and the other detector without MPM. All detectors are trained and tested on the same adversarial method.

Carlini \& Wagner \cite{carlini2017adversarial} point out that it is necessary to evaluate defenses using a strong attack on harder datasets (such as Imagenet). Moreover, Carlini \& Wagner prove that using a second neural network to identify adversarial examples is the least effective defense. Therefore we only compare our method with the defense which is effective for C\&W on Imagenet and not based on another neural network. However, Li's defense \cite{li2017adversarial} fails against the C\&W attack. The Hendrycks defense \cite{hendrycks2016early} is only effective for MNIST. Grosse's \cite{grosse2017statistical}, Gong's \cite{gong2017adversarial} and Metzen's \cite{metzen2017detecting} defenses use a second neural network to classify images as normal or adversarial. Lu's defense \cite{Lu2017SafetyNet} has good performance on Imagenet even though its performance against C\&W is not evaluated. Finally we compare our detectors with Lu's defense which is denoted as RBF-SVM.

\begin{figure}[htbp]
  \centering
  \includegraphics[width=0.5\textwidth]{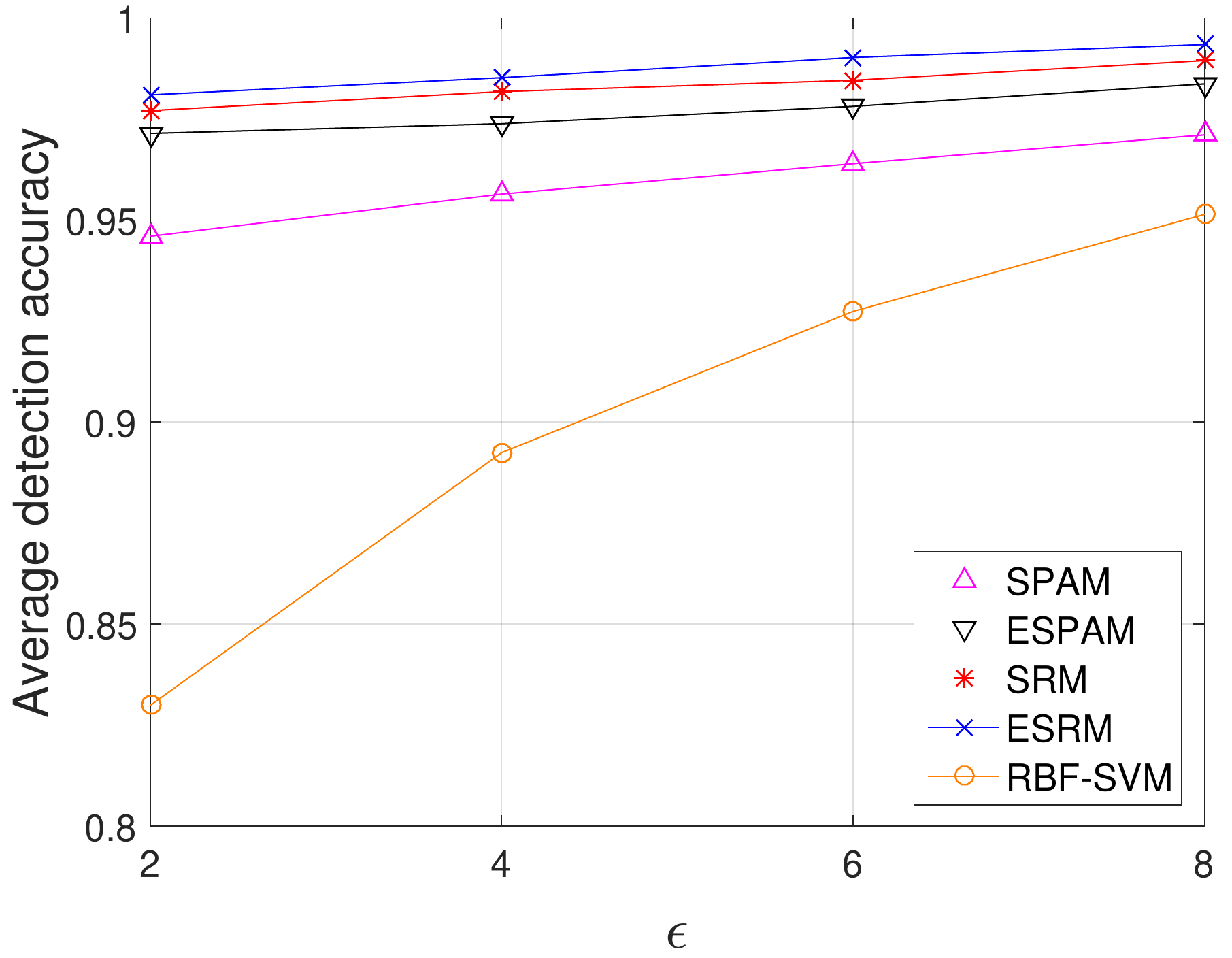}
  \caption{Average detection accuracy for detectors against FGSM.}
  \label{figure1}
\end{figure}

\begin{figure}[htbp]
  \centering
  \includegraphics[width=0.5\textwidth]{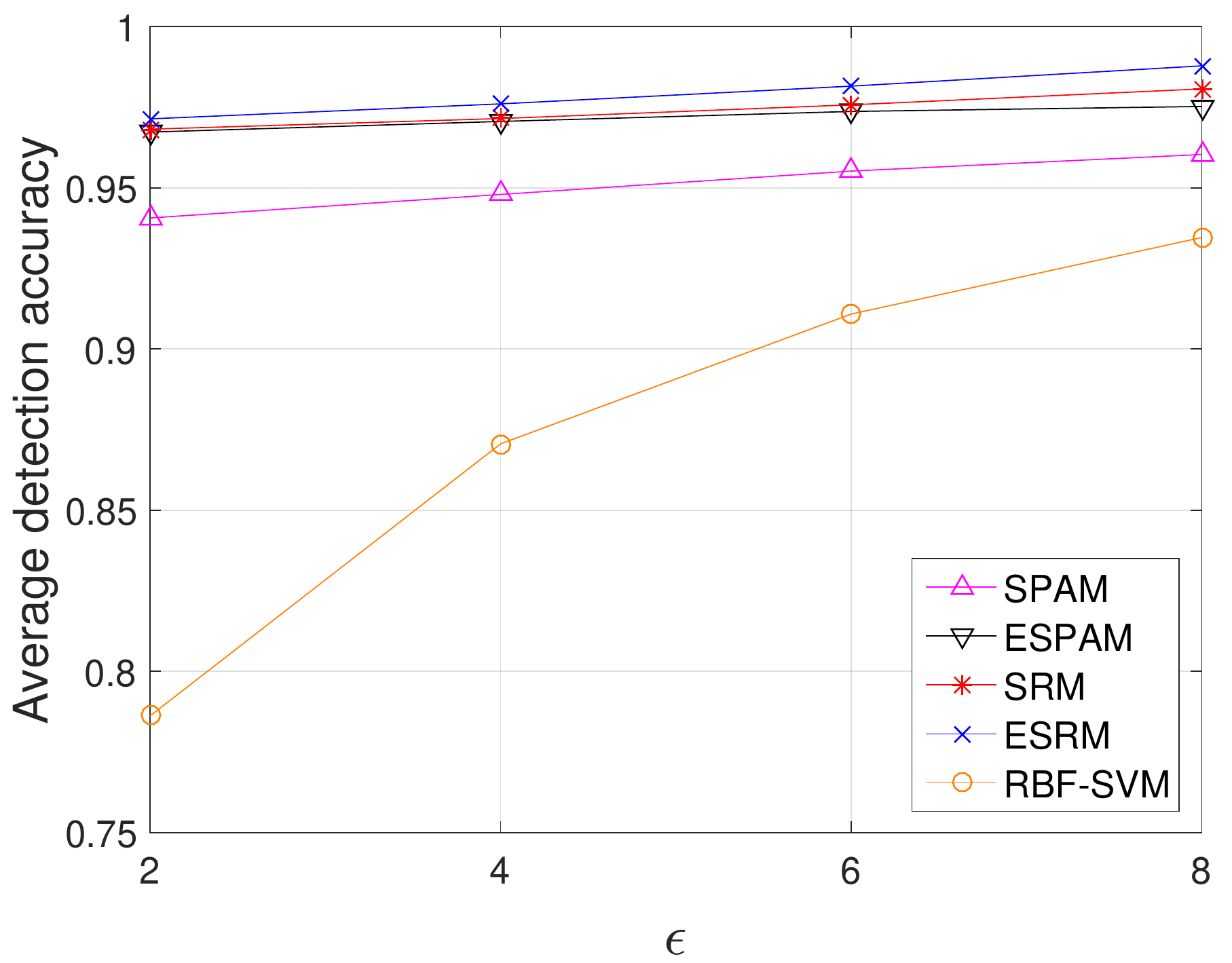}
  \caption{Average detection accuracy for detectors against IGSM.}
  \label{figure2}
\end{figure}

The experimental results of detecting adversarial examples are shown in Table $1-4$. The data of Table $1-4$ is the detection accuracy of normal images and adversarial images. Figure \ref{figure1} and Figure \ref{figure2} illustrate these detectors' performance by averaging the accuracy of detecting normal images and adversarial images. First of all, the results reveal that the detectors with MPM have higher detection accuracy. Moreover, MPM has stronger enhancing effect on SPAM than SRM. When detecting FGSM and IGSM, ESPAM even has comparable performance as SRM. That is to say, we can even use the low-dimensional model to achieve comparable performance as the high-dimensional model via the enhancing method. Experimental results show that it is difficult to detect adversarial examples generated by the C\&W method. RBF-SVM is almost invalid against C\&W. SPAM and SRM achieve relatively low accuracy when detecting C\&W. However, MPM improves SPAM by more than 15 percent and the detection accuracy of ESRM reaches 93 percent on detecting adversarial examples yielded by C\&W. In addition, the detection accuracy of ESRM is the highest when detecting FGSM, IGSM, Deepfool and C\&W. However, the computation time of SRM and ESRM is much longer because of their high-dimensional features.

%According to the results of the experiment, our detecting method can correctly distinguish normal images and adversarial images when the detector is trained and tested on the same adversarial method in most cases. However, the performance of our detection model is not good enough against C\&W. In addition, the detector trained for $\epsilon=2$ generalizes well to large $\epsilon$ when detecting adversarial examples generated by fast gradient sign method and iterative gradient sign method.

\section{Conclusions}

Inspired by the insight of Goodfellow et al. \cite{Goodfellow2015Explaining} that ``adversarial examples can be thought of as a sort of accidental steganography'', we propose to apply steganalysis to detecting adversarial examples. We also propose a method to enhance steganalysis features. The experimental results show that the enhanced scheme can accurately detect various kinds of adversarial attacks including the C\&W method. Moreover, the secondary adversarial attacks \cite{carlini2017adversarial} cannot be directly performed to our method because the structure of our detection model is not a neural network. Therefore an open problem is how to implement secondary attacks on our proposed defense method. %However, our detector cannot have very good performance when it is not trained and tested on the same adversarial method. In future work, we will try to explore methods for constructing a set of detectors against different kinds of adversarial attacks and performing parallel detection.%In future work, we will try to explore methods for constructing one detector against different kinds of adversarial attacks.
%In future work, we will try to explore methods for constructing a set of detectors against different kinds of adversarial attacks and performing parallel detection.

\bibliography{sample-bibliography}
\bibliographystyle{aaai}

\end{document}